\documentclass[anonymous=false, %
               format=acmsmall, %
               review=false, %
               screen=true, %
               nonacm=true,
               ruled,
               vlined]{acmart}

\usepackage{hyperref}
\usepackage{url}
\usepackage{amsmath}
\usepackage{graphicx}
\usepackage{subcaption}
\usepackage[ruled,vlined]{algorithm2e}
\usepackage{tikz}
\usetikzlibrary{positioning}
\usetikzlibrary{quotes,arrows.meta}
\usetikzlibrary{positioning}
\usepackage{array, multirow}
\begin{document}

\renewcommand{\arraystretch}{1.2}
\urlstyle{tt}
\citestyle{acmauthoryear}

\title{Compartmental Models for COVID-19 and Control via Policy Interventions}

%  \titlenote{This is a titlenote}
%  \subtitle{This is a subtitle}
%  \subtitlenote{Subtitle note}

\author{Swapneel Mehta}
\affiliation{%
  \institution{New York University}
  \department{Center for Data Science}
}
\email{swapneel.mehta@nyu.edu}

\author{Noah Kasmanoff}
\affiliation{%
  \institution{New York University}
  \department{Center for Data Science}
}
\email{nsk367@nyu.edu}

% \author{Jean-Baptiste Tristan}
% %\orcid{1234-5678-9012-3456}
% \affiliation{%
%   \institution{Boston College}
%   %\city{Chestnut Hill}
%   %\state{MA}
%   %\postcode{02467}
%   %\country{USA}
% }
% \email{jean.baptiste.tristan@oracle.com}

% \author{Jan-Willem van de Meent}
% %\orcid{1234-5678-9012-3456}
% \affiliation{%
%   \institution{Northeastern University}
%   %\department{Khoury College of Computer Sciences}
%   %\streetaddress{360 Huntington Ave}
%   \city{Boston}
%   \state{MA}
%   %\postcode{02115}
%   \country{USA}
% }
% \email{j.vandemeent@northeastern.edu}
%\renewcommand\shortauthors{Mage, M. et al}

\begin{abstract}
  We demonstrate an approach to replicate and forecast the spread of the SARS-CoV-2 (COVID-19) pandemic using the toolkit of probabilistic programming languages (PPLs). Our goal is to study the impact of various modeling assumptions and motivate policy interventions enacted to limit the spread of infectious diseases. Using existing compartmental models we show how to use inference in PPLs to obtain posterior estimates for disease parameters. We improve popular existing models to reflect practical considerations such as the under-reporting of the true number of COVID-19 cases and motivate the need to model policy interventions for real-world data.  We design an SEI3RD model as a reusable template and demonstrate its flexibility in comparison to other models. We also provide a greedy algorithm that selects the optimal series of policy interventions that are likely to control the infected population subject to provided constraints. We work within a simple, modular, and reproducible framework to enable immediate cross-domain access to the state-of-the-art in probabilistic inference with emphasis on policy interventions. \textbf{We are not epidemiologists}; the sole aim of this study is to serve as a exposition of methods, not to directly infer the real-world impact of policy-making for COVID-19.

\end{abstract}

\maketitle

\section{Introduction}

In order to understand and control infectious diseases, it is important to build realistic models capable of accurately replicating and projecting their transmission in a region \cite{atkeson2020will, tang2020mathematic, sameni2020mathematical}. The underlying assumption being that a model capable of replicating disease spread has captured the true causal dynamics sufficiently well. Motivated by this, there has been a spate of research in applying SIR, SEIR, and SEI3RD models to this problem \cite{covid2020modeling, lopez2020end}. These belong to a class of compartmental models that are underpinned by Lotka-Volterra dynamics that divide a population into sections or compartments and describe the probabilistic transitions between them through a set of partial differential equations. We extend this direction of work with an emphasis on modeling under-reported cases, decoupling policy interventions from compartmental transitions, estimating the impact of policy interventions, selecting a sequence of optimal interventions to control the spread of diseases, and using the SEI3RD variant as an extension of existing work on SEIR models since it forms a template for many other extensions of compartmental models \cite{giordano2020modelling, senapati2020impact, kennedy2020modeling, wolff2020build, grimm2021extensions, winters2020novel}. The authors of \cite{hong2020estimation}, much like us, propose a new statistical tool to visualize analyses of COVID-19 data. However, our framework makes it much simpler to inspect the intricacies of modeling assumptions, expand with a custom set of constraints, and experiment with counterfactual simulations without the need for extensive compute or data. We have created some demo notebooks which will be made available publicly \footnote{\url{https://drive.google.com/drive/folders/1Npdn4bS_qlps5EdA6vXlMvRXSljGgYCd?usp=sharing}}.

In light of probabilistic programming languages (PPLs) \cite{bingham2019pyro, van2018introduction, salvatier2016probabilistic, carpenter2017stan} reaching their 'coming-of-age' moment, COVID-19 modeling is being explored to guide and support policy decisions and decision-makers \cite{wood2020planning, de2020simulation}. In this work, we provide data scientists with a concrete example of applying probabilistic inference to understand disease spread and control. To epidemiologists, we offer this manuscript as a guide to design compartmental models to evaluate the impact of policy interventions \cite{mandal2020prudent, giordano2020modelling, wang2020epidemiological}. We include clear motivation and detailed real-world examples of how to manually explore the impact of such non-pharmaceutical interventions (NPIs) using synthetically generated data and provide an algorithm to automatically generate strategies for implementing governmental policies for disease control.

% In section 2, we briefly highlight the problem setup and motivation in the context of related work. Section 3 discusses the model dynamics and lists the assumptions made in our work. In section 4, we offer a brief technical walk-through of our work. Sections 5 and 6 discuss the novel approach that improves our modeling assumptions. Our results are briefly summarised in section 7. In section 8, we describe an adaptive technique of picking policy interventions.

Our main contributions are as follows:
\begin{itemize}
    \item Improve existing compartmental models with the ability to deal with the under-reporting of COVID-19 cases and decoupling NPIs from disease parameters.
    \item Evaluate our parameter estimates through empirical comparisons with those in prevalent literature, observed patterns in testing coverage, and highlight the relative consistency of our predictions compared to anomalies in existing approaches.
    \item Model fixed policy interventions and describe an algorithm to select adaptive policy interventions to aid government efforts to limit the spread of disease.
    \item Highlight the ease of using PPLs to design, extend, and fit SEI3RD models using approximate inference to obtain posterior disease parameter estimates; in addition to having an open-source code-base and our self-contained tutorial to be released at the time of publication.
\end{itemize}

\section{Susceptible-Infected-Recovered (SIR) Model Dynamics}

The SIR model dynamics form a template for the increasingly complex compartmental models we explore. These dynamics are encapsulated in the following differential equations following the Lotka-Volterra Model:

\begin{equation*}
    \frac{dS}{dt} = -\beta S(t) i(t) \quad , \qquad
    \frac{dI}{dt} = \beta S(t) i(t) - \gamma I(t) \quad , \qquad
    \frac{dR}{dt} = \gamma I(t) 
\end{equation*}

\begin{figure}
    \centering
    
    \begin{tikzpicture}[%
      every node/.style={draw,rectangle,rounded corners, minimum size=0.8cm},node distance=2cm]
      % the vertices
      \node[fill=yellow!25] (source) {S};
      \node[fill=red!20,right=of source] (three) {I};
      \node[fill=green!20,right=of three] (six) {R};
      % the edges
      \draw [->] (source) edge (three) -- (three) node [midway,above=-5pt, fill=none,draw=none] {$\beta$} (three) edge (six) -- (six) node [midway,above=-2pt, fill=none,draw=none]{$\gamma$} ;
    \end{tikzpicture}

    \caption{The SIR Compartmental Model}
    \label{fig:sir_model}
\end{figure}
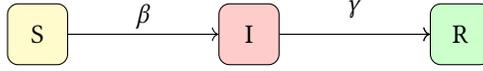
    
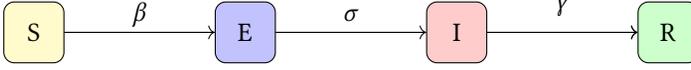
\begin{figure}
    \centering
    
    \begin{tikzpicture}[%
      every node/.style={draw,rectangle, rounded corners, minimum size=0.8cm},node distance=2cm]
      % the vertices
      \node[fill=yellow!25] (source) {S};
      \node[fill=blue!24,right=of source] (three) {E};
      \node[fill=red!20,right=of three] (four) {I};
      \node[fill=green!20,right=of four] (six) {R};
      % the edges
      \draw [->] (source) edge (three) -- (three) node [midway,above=-5pt, fill=none,draw=none] {$\beta$} (three) edge (four) -- (four) node [midway,above=-5pt, fill=none,draw=none] {$\sigma$} (four) edge (six) -- (six) node [midway,above=-2pt, fill=none,draw=none]{$\gamma$} ;
    \end{tikzpicture}

    \caption{Add an 'Exposed' compartment to the SIR model to obtain the SEIR model}
    \label{fig:seir_model}
\end{figure}

\section{Modeling Noisy Observations}

The population-level disease parameters we discussed for the SIR model (transmission probability $\beta$ and recovery rate $\gamma$) are usually estimated in terms of measurable quantities from observed data such as the mean recovery time. The response rate $\rho$ indicates the proportion of observed infections in reality because we typically cannot expect to observe every single case of infection. To be clear, it is often the case that people may not realize they have been infected in the absence of extensive testing. This has been the case in most countries at least for the initial few months of the pandemic and some strategies attempt to remedy it in different ways \cite{khan2021under, lachmann2020correcting, jagodnik2020correcting}. Since the real-world data we intend to use overlaps significantly with this duration, it makes sense to add this variable to our model. Importantly, the same may be exacerbated due to socioeconomic and political factors, so $\rho$ allows us to model this under-reporting of COVID-19 cases in practice. We start by making a guess about the range for the reproduction number $R_0 = \dfrac{\beta}{\gamma}$ and response rate $\rho$. This 'guess' is equivalent to placing empirically informed (from past outbreaks, for instance) priors on these latent variables and estimate their posterior distributions.

\subsection{Using PPLs for Modeling and Inference}

PPLs are a natural candidate among the tools we considered for this problem. We would like to define a time-series probabilistic model with some underlying dynamics that encodes our assumptions about the data-generating process; then we want to fit this model to the data through stochastic variational inference \cite{hoffman2013stochastic, ranganath2014black}, and use the posterior distributions over its parameters to make predictions about future time-steps. As a sanity check, we also want to test how well it replicates the historical trajectory of the disease through simulations. \texttt{Pyro} \cite{bingham2019pyro}, a deep universal probabilistic programming language, is an excellent candidate for this because of the following reasons:

\begin{itemize}
    \item It comprises of thin wrappers around \texttt{PyTorch} \cite{paszke2019pytorch} distributions allowing us to write complex generative models interweaving stochastic and deterministic control flow while still within a familiar and popular machine learning (ML) framework.
    \item It offers general-purpose inference algorithms out of the box that allows us to shift the focus from designing custom inference algorithms to building expressive models.
    \item It recently extended support in terms of an API for epidemiological models. We perform our experiments in \texttt{Pyro} since it is a much more accessible framework due to the community around it in comparison to modern alternatives like \texttt{Pyprob} \cite{baydin2019etalumis}. We can confirm this empirically since we spent some time reproducing work along similar lines in \cite{wood2020planning} and found much less time required to do the same in \texttt{Pyro}. We do still appreciate the relative ease of reproducing it compared to other machine learning research for healthcare.
\end{itemize}

We consider a partially observed population and want to understand the putative controls that could achieve a goal defined as "control the spread of the disease", "reduce the death rate", or other such equivalent outcomes. While some of the parameters that define disease spread are controllable, there are certain non-controllable parameters including properties of the disease that we will infer. From a computational standpoint, we demonstrate the capability of the universal probabilistic programming language, \texttt{Pyro}, to combine a system of equations that define a simulator and perform inference over the latent variables within the simulation to obtain a posterior distribution over the model parameters. In addition to an inference engine, \texttt{Pyro} offers the capability to intervene on the variables within this simulation in order to obtain potential outcomes of policy changes that can be expressed within the language as fixing the values of certain stochastic parameters. 
\begin{figure*}[ht]
	\begin{subfigure}{.45\textwidth}
		\centering
		% include first image
		\includegraphics[width=6.3cm, height=3.5cm]{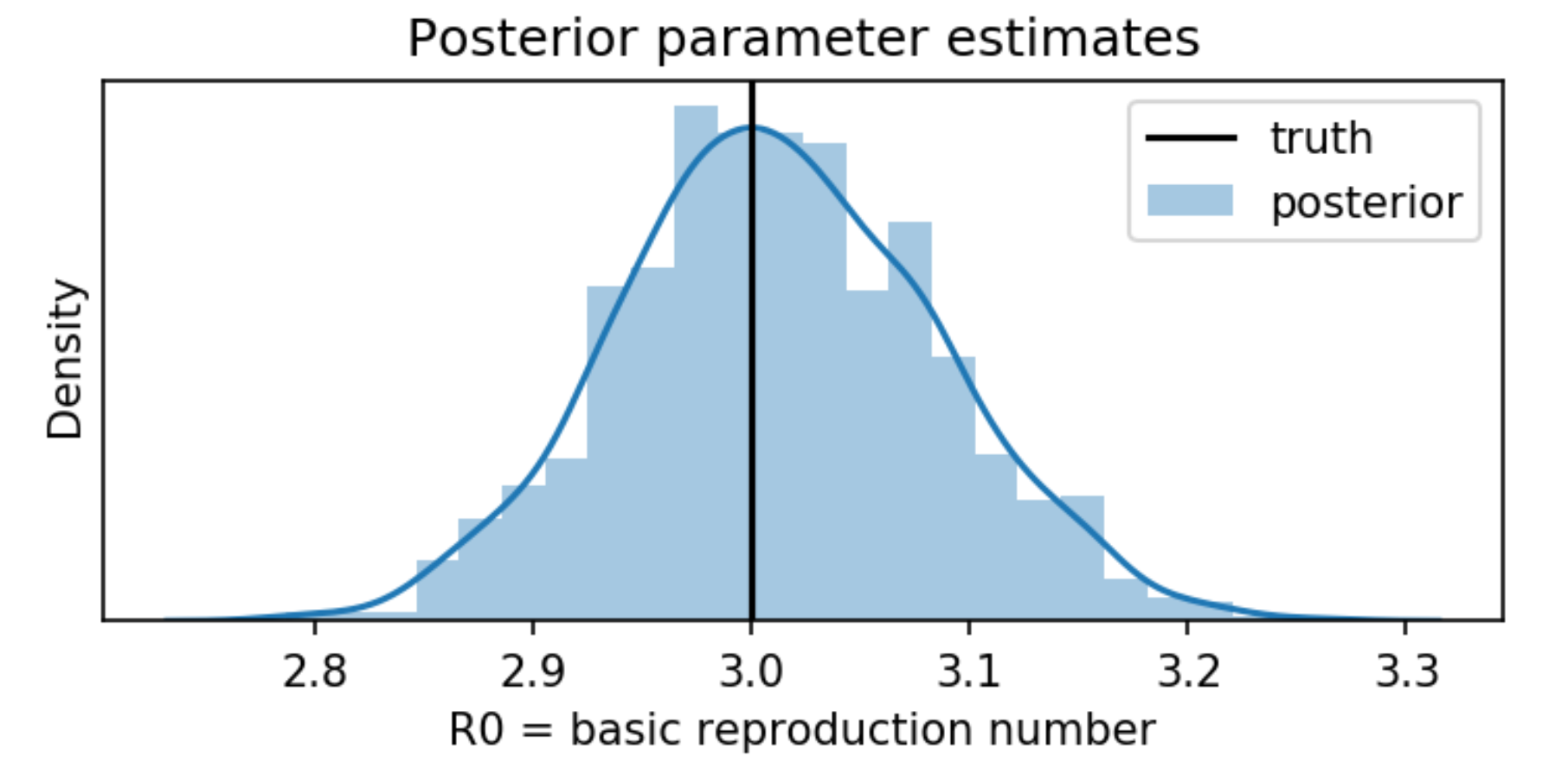}
		\caption{Posterior distribution for $R0$}
		\label{fig:posterior_R0}
	\end{subfigure}
	\begin{subfigure}{.45\textwidth}
		\centering
		% include second image
		\includegraphics[width=6.3cm, height=3.5cm]{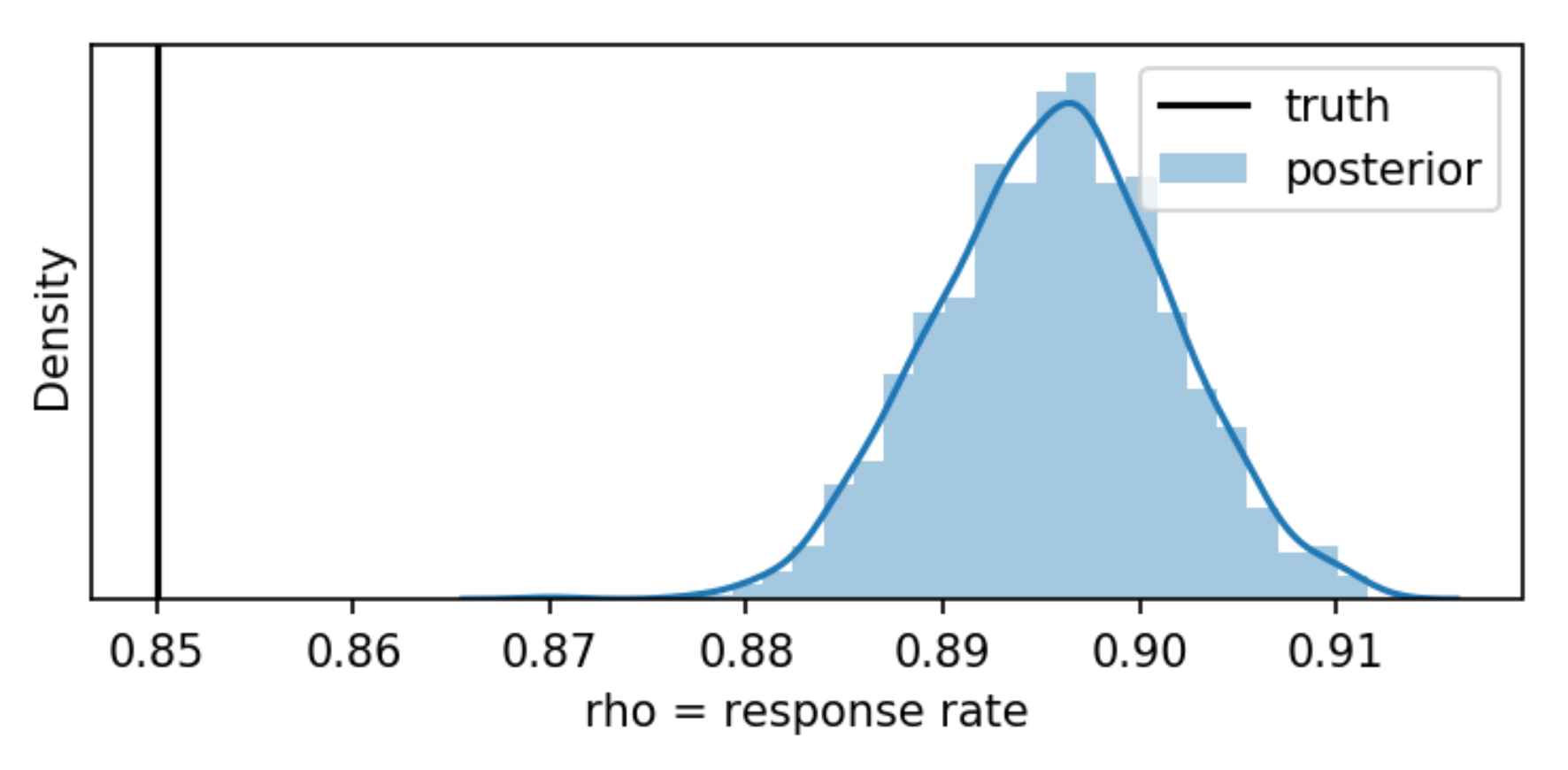}
		\caption{Posterior distribution for $\rho$}
		\label{fig:posterior_rho}
	\end{subfigure}
	\label{fig:posteriors}
\end{figure*}

\begin{figure*}[ht]
	\begin{subfigure}{\textwidth}
		\centering
		% include first image
		\includegraphics[width=12cm, height=8cm]{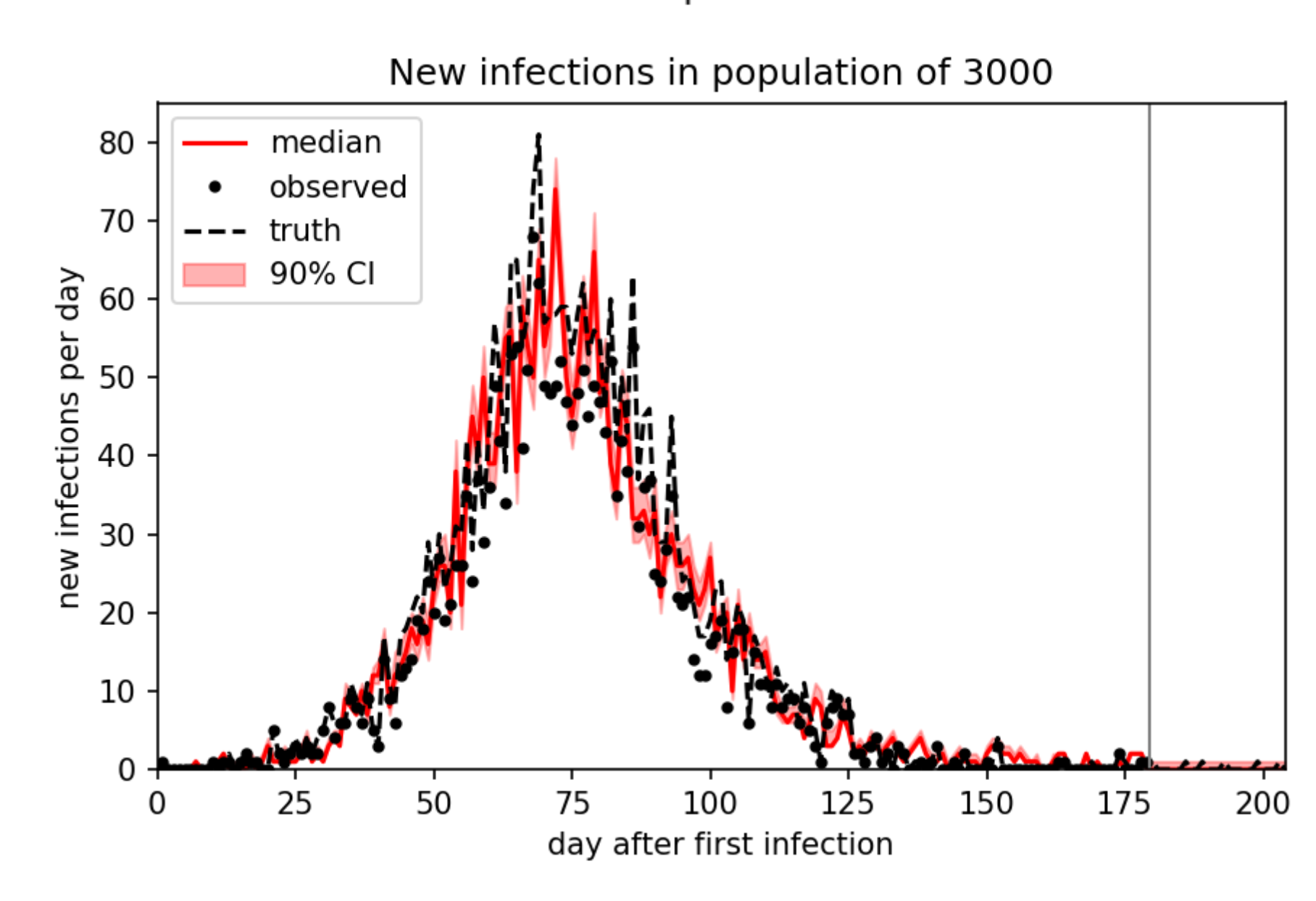}
	\end{subfigure}
	\caption{Predicting Future Infections for a Simulated Dataset in a Highly Infectious Setting}
	\label{fig:forecasting}
\end{figure*}

\section{Experimental Evaluation of Compartmental Models in Real-world Settings}

For our simulation-based study, we used models $M = {SIR, SEIR, SEI3RD}$ and refined versions of each, $M(i)$,
wherein we start with an infected population of 0.01\%. We conduct some simulation studies for instance by setting the $R_0 = 3.0$ and $\rho = 0.85$. We perform variational inference \cite{hoffman2013stochastic, ranganath2014black} to obtain posterior parameter estimates close to the true values. These are indicated by the vertical black line in \ref{fig:posterior_R0} and \ref{fig:posterior_rho}. Since the posterior point-estimates are close to the true values for the simulation, we can see why we are able to accurately replicate the disease spread shown in \ref{fig:forecasting}. We plot the daily infections versus time and the plot resembles a single wave of infected patients. The shaded portion indicates a $90\%$ confidence interval for the model's predictions which closely tracks the true values of disease spread. While it is no surprise that SIR, SEIR, and SEI3RD models perform well on simulated data, there are extensive studies in applying the SEIR model in practice. However, we believe that the SEIR model also has its failures (see \nameref{section:qualitativeanalysis}) that often go undetected due to a lack of comprehensive evaluation. We show this through an extensive set of experiments on real-world data across multiple geographies. The tabular comparison of the reproduction number or $R0$ and the $\rho$ estimated via our models are shown in Table \ref{table:covid_R0} for different time periods which offers the following insights:

\begin{itemize}
    \item The reported cases for the initial period of January - July 2020 indicate far less testing than the full periods (upto May 2021 for the USA and January 2021 for the rest of the world, in our dataset).
    \item The estimated $R0$ seemed high for most regions upto April 2020, and even though longer-term estimates denoted an $R0$ of a little over 1 for most regions, the spate of cases underscores the need to reduce the spread of the infection. At the same time, the reproduction number is not all that matters and a careful study of mortality rates is warranted as testing increases and fatalities decrease in order to draw concrete scientific conclusions and policy recommendations given this evidence.
    \item We show qualitatively why it is necessary to build more granular, flexible compartmental models and underscore the need for separately modeling policy interventions as human influence on these processes.
\end{itemize}

The references for \nameref{table:covid_R0} are drawn from prevalent literature and expert-curated resources\footnote{\url{https://epiforecasts.io/covid/posts/national/united-states/}} \footnote{\url{https://covidestim.org/us/GA}} on the subject \cite{kamalichtime, lau2020characterizing, prodanov2021analytical, Gunzler2020}. A certain George E. P. Box would be wont to say "All $R0$ estimates are wrong but some may be useful", and we illustrate this via an empirical and rather qualitative route. Firstly, in some cases the reference method itself presents anomalous forecasts like the $R0$ for Germany throughout the year being predicted as an unusual 22.032, and that of the Netherlands, 9.103 from \cite{prodanov2021analytical}. In comparison, we observe relative consistency at least at the model level, which might be a strong signal to consider more realistic modeling choices when using real, noisy data for estimation. Where we lack consistency, we have a signal in the form of a set of forecasts as well as an additional estimate of $\rho$. An oddly low $\rho$, for instance, might convince us of an estimation error despite having high-confidence and a good fit to the observed data. We observe that the SEI3RD models perform better than the SEIR and SIR models which might lack the capacity to effectively model transitions. We also study a refined set of compartmental models which consider an initial, partially infected (0.01\%) population instead of starting from a single infection ('patient zero').

Once we successfully model disease progression and forecast for a set of future time-steps to recover potential new infections. We can examine counterfactual questions of the nature 'What would have happened to the number of cases if the government enacted $X$ steps at $Y$ time?' by introducing policy interventions $u$ at this stage and figure out the global minimally invasive intervention to perform that will remain within the desired thresholds for the infected populace. An adaptive algorithm to determine the optimal policy interventions is developed in this paper as we continue to explore breadth-wise analyses pertaining to COVID-19 control, and compare our models to the impact of real-world interventions. This discussion, albeit a core contribution, is relegated to the appendix.

\section{The Reproduction Number}

As responsible data scientists attempting to provide a tool for epidemiological analysis, it is important not to overstate the relevance of estimating the correct $R0$ from the data since that is not the only parameter of interest. Many numbers have been bandied about in the news under the assumption that controlling $R0$ implies controlling the pandemic. To a certain degree this is true, however there are important caveats to this notion that some have expanded upon \cite{HebertDufresne2020, maruotti2021misuse}. The summary of the discussion is that $R0$ must be used as a tool to paint a partial picture of a disease spread, in conjunction with multiple factors. In this regard, our consideration of $\rho$, modeling of infected fractions of populations, and consideration of policy interventions are significant steps taken to provide a comprehensive idea of the state of a pandemic.

\begin{table*}
  \caption{\label{table:covid_R0} Comparing $R0$ estimates with literature}
  \begin{tabular}{c | l | lll | lll}
     \toprule
    Region &Model &\multicolumn{3}{l}{  Jan - April, '20}  &\multicolumn{3}{l}{  Jan - Dec, '20} \\
    \hline
       & &$R0$ &$\rho$ &Ref. $R0$  &$R0$ &$\rho$ &Ref. $R0$ \\
    \midrule
     \multirow{6}{*}{Italy} &SIR &1.48 $\sigma=.01$ &0.187 &2.676 &1.34 &0.125 &1.8024\\
      &SIR(i) &1.712 &0.76 $\sigma=.03$ &- &0.608 &0.506 &-\\ \cline{2-8}
      &SEIR &2.27 $\sigma=.02$ &0.11 &- &1.49 &0.14 &-\\
      &SEIR(i) &0.403 &0.56 &- &1.96 &0.178 &-\\ \cline{2-8}
      &SEI3RD &3.03 $\sigma=.07$ &0.557 &- &5.78 &0.001 &-\\
      &SEI3RD(i) &1.247 &0.532 &- &0.215 &0.505 &-\\
     \hline
     \multirow{6}{*}{Netherlands} &SIR &1.69 &0.36 $\sigma=.01$ &1.9962 &1.724 $\sigma=0.84$ &0.71 &9.103\\
      &SIR(i) &1.699 $\sigma=.01$ &0.54 $\sigma=.015$ &- &0.490 &0.5079 &-\\ \cline{2-8}
      &SEIR &4.51 $\sigma=.011$ &3.24 $\sigma=.01$ &- &1.76 &0.45 &-\\
      &SEIR(i) &10.048 &0.71 &- &0.236 &0.50 &-\\ \cline{2-8}
      &SEI3RD &4.23 $\sigma=.04$ &0.735 &- &2.78 &0.154 &-\\
      &SEI3RD(i) &3.763 $\sigma=.03$ &0.70 &- &3.51 &0.108 &-\\
     \hline
     \multirow{6}{*}{New York} &SIR &1.432 &0.54 $\sigma=.016$ &1.21 &1.475 &0.027 &0.81\\
      &SIR(i) &1.643 &0.594 &- &2.078 &0.619 &-\\ \cline{2-8}
      &SEIR &0.88 &0.50 &- &0.81 &0.50 &-\\
      &SEIR(i) &4.09 $\sigma=.01$ &0.04 &- &4.98 &0.14 &-\\ \cline{2-8}
      &SEI3RD &1.53 &0.504 &- &1.67 &0.503 &-\\
      &SEI3RD(i) &10.79 $\sigma=.006$ &0.0795 &- &12.295 &0.253 &-\\
     \hline
     \multirow{6}{*}{Germany} &SIR &0.108 &0.5186 &1.637 &1.610 &0.120 &22.032\\
      &SIR(i) &1.643 &0.594 &- &2.078 &0.619 &-\\ \cline{2-8}
      &SEIR &2.51 $\sigma=.013$ &0.76 $\sigma=.01$ &- &2.93 $\sigma=.1$ &0.154 &-\\
      &SEIR(i) &1.973 &0.213 &- &5.02 $\sigma=.47$ &0.57 &-\\ \cline{2-8}
      &SEI3RD &0.572 &0.55 &- &5.56 $\sigma=.08$ &0.01 &-\\
      &SEI3RD(i) &1.37 &0.52 &- &4.7 &0.124 &-\\
     \hline
     \multirow{6}{*}{Georgia} &SIR &3.74 &0.303 &1.45 &3.296 &0.343 &0.84\\
      &SIR(i) &4.075 &0.302 &- &3.661 &0.039 &-\\ \cline{2-8}
      &SEIR &5.3 $\sigma=.17$ &0.015 &- &3.47 &0.56 &-\\
      &SEIR(i) &4.97 $\sigma=.078$ &0.014 &- &5.36 &0.096 &-\\ \cline{2-8}
      &SEI3RD &4.09 $\sigma=.17$ &0.5483 &- &1.32 &0.501 &-\\
      &SEI3RD(i) &6.8 &0.049 &- &5.62 &0.205 &-\\
   \bottomrule
  \end{tabular}
\end{table*}

\begin{acks}
We would like to acknowledge guidance and support from Kyle Cranmer and Rajesh Ranganath that laid the foundations for this research project.
\end{acks}

\newpage

\bibliographystyle{acm-reference-format}
\bibliography{probprog-2021-instructions}

\newpage

\appendix

\section{Appendix}

% \begin{table}
%   \label{table:refined_covid_R0}
%   \caption{$R0$ Estimates from a modified SEIR Model with an initial infected fraction of population}
%   \begin{tabular}{c | ll | ll}
%      \toprule
%      &\multicolumn{2}{l}{01/20 - 04/20}  &\multicolumn{2}{l}{01/20 - 01/21} \\
%     \hline
%       Region  &$R0$ &$\rho$  &$R0$ &$\rho$ \\
%     \midrule
%      Netherlands \\
%      Italy \\
%      Germany \\
%      New York \\
%      Georgia \\
%   \bottomrule
%   \end{tabular}
% \end{table}

\section{Qualitative Analysis of Successes and Failures}
\label{section:qualitativeanalysis}

We have attempted to fit many different kinds of models and obtained certain parameter estimates which we compare to other, equally noisy parameter estimates from the literature. The fact is, each country has dealt with COVID-19 in different ways. While China seemed to be able to take stringent measures and limit the spread, others like Italy were not able to follow the same approach. There was a strong focus on the fast-rising numbers of infected people in the United States (September, 2020), quickly overshadowed by a worse state of affairs in India (May, 2021). It is crucial to analyse how well each model is able to not only fit the data but also forecast the spread of disease accurately. Since we have trained some models on partial amounts of data, let us take a look at which of the models can identify upcoming 'waves' in the progress of the disease.

\begin{figure*}[ht]
	\begin{subfigure}{\textwidth}
		\centering
		% include first image
		\includegraphics[width=13cm, height=8cm]{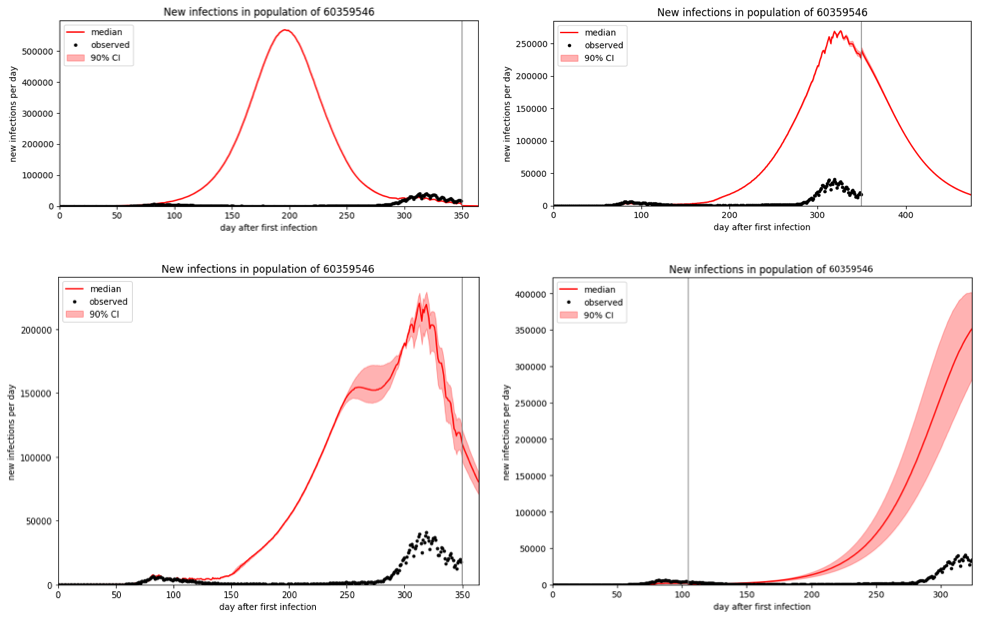}
	\end{subfigure}
	\caption{Fitting SIR (top left), SEI3RD (bottom right) and SEIR models to data from Italy}
	\label{fig:itaquals}
\end{figure*}

\begin{figure*}[ht]
	\begin{subfigure}{\textwidth}
		\centering
		% include first image
		\includegraphics[width=13cm, height=7cm]{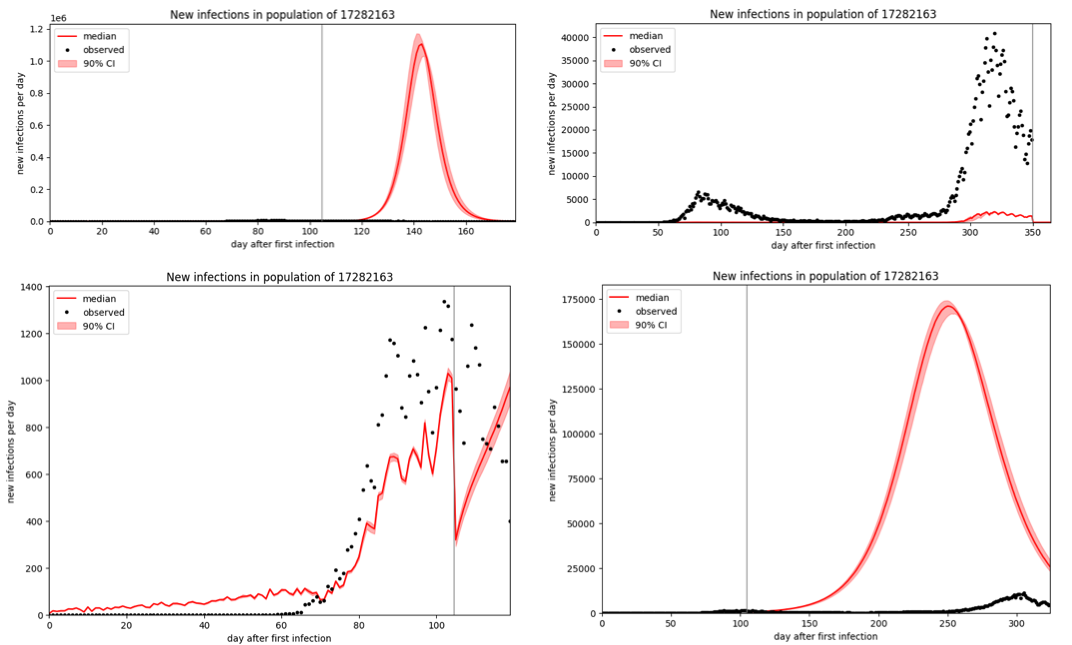}
	\end{subfigure}
	\caption{Fitting SIR (top left), SEIR (top right) and SEI3RD models to data from the Netherlands}
	\label{fig:nldquals}
\end{figure*}

\begin{figure*}[ht]
	\begin{subfigure}{\textwidth}
		\centering
		% include first image
		\includegraphics[width=13cm, height=4cm]{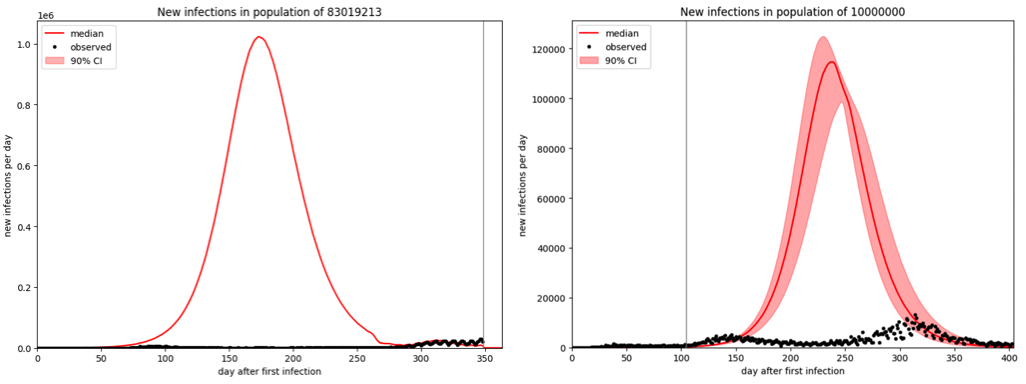}
	\end{subfigure}
	\caption{Fitting SEI3RD models to data from Germany and Georgia (right)}
	\label{fig:deugaquals}
\end{figure*}

For instance, in the figures \ref{fig:itaquals} (vertical bars mark the beginning of forecasts), the SEIR and the SEI3RD models are able to capture the second wave, with the latter being able to, impressively enough, predict the second one having only observed the first wave. Similarly, in \ref{fig:nldquals} we can see that both the SIR and SEIR models seem to perform poorly at data fitting and forecasting. Note that in both cases, the notion of the 'best model' as indicated by comparing estimated $R0$ does not seem to hold, emphasising the need to consider the least restrictive models when conducting such studies. Of course, the SEI3RD model is not a panacea. We offer two additional examples of its forecasts \ref{fig:deugaquals} that convince us we need to model an external factor that allows us to modify the second, larger wave (peak) in a manner similar to how government interventions delayed the infections from successively peaking as per the predictions of the SEI3RD model. Also in \ref{fig:deugaquals} is the availability of confidence intervals for the forecast. In general, while the SEI3RD model performs well at capturing multiple waves of COVID-19 infections, we are making a fundamentally incorrect modeling assumption by ignoring the fact that disease evolution was hindered by human intervention, in particular via government policies designed to limit its spread. In fact, we can utilise these 'vanilla' SEI3RD models that are reasonably good at predicting the potential spread of a disease to help us figure out how best to limit it! This motivates us to study \nameref{section:policy_interventions}.

\begin{figure*}[ht]
	\begin{subfigure}{\textwidth}
		\centering
		% include first image
		\includegraphics[width=13cm, height=7cm]{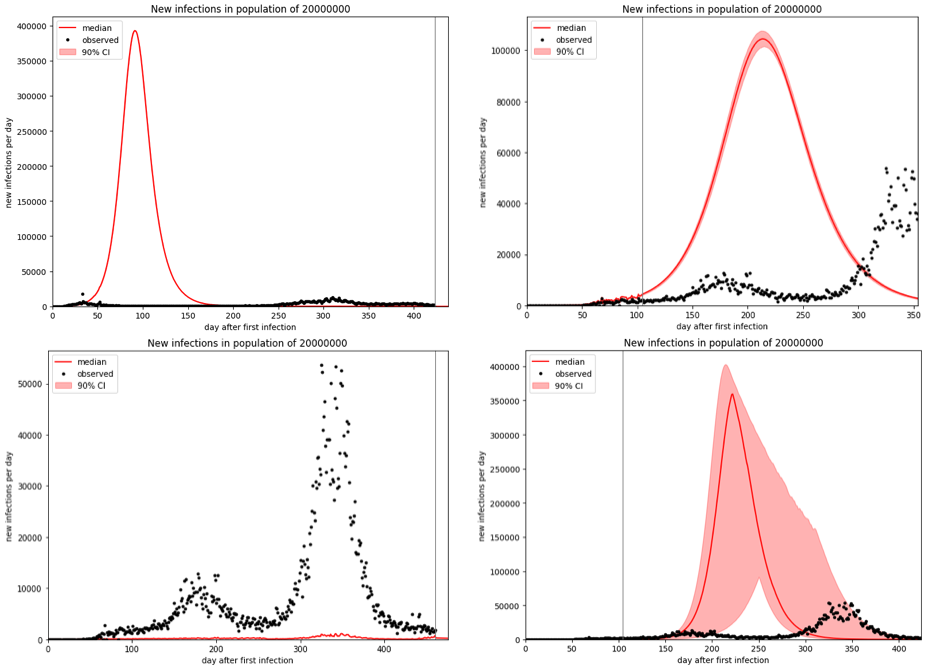}
	\end{subfigure}
	\caption{Fitting SIR (top left), SEIR (top right) and SEI3RD models to data from New York}
	\label{fig:nycquals}
\end{figure*}

\section{Compartmental Model Extensions in Pyro}
\label{section:extensions}

There is a vast body of literature on compartmental models for epidemiology. In particular, the SEIR model seems the most popular and widely applied for real-world correspondence possibly owing to a trade-off between simplicity and effectiveness \cite{buckman2020replicating}. In \cite{covid2020modeling} the team fits an SEIR model to mortality data in an effort to examine possible trajectories of COVID-19 infections at the state level. In particular, they conduct a review of COVID-19 model on a state-wise basis with particular emphasis on charting out potential scenarios in terms of the number of fatalities in the presence of various non-pharmaceutical interventions. They forecast the best and worst case outcomes in terms of these numbers and make recommendations to ensure the safety of the US population in case of epidemic resurgences in many states. The authors of \cite{lopez2020end} use an SEIR model to conduct a study of the recurrence of the COVID-19 pandemic via different post-confinement scenarios. Their work highlights the importance of non-pharmaceutical interventions due to the re-emergence risk from the time-decay of acquired immunity and lack of effective pharmaceutical interventions.

There is however significant work that has gone into building more expressive, realistic, informed models some of which are SIDARTHE \cite{giordano2020modelling}, SEI3HR in India \cite{senapati2020impact}, SUEIHCDR in Brazil \cite{kennedy2020modeling}, SEI3RSD \cite{wolff2020build}, SEI3Q3RD \cite{grimm2021extensions}, SEI3R2S \cite{winters2020novel}, and others \cite{ndairou2020mathematical}. SEIR models for the spread of disease \cite{buckman2020replicating, covid2020modeling, buckman2020replicating} are prevalent in the existing literature with multitudinous global, national, and regional studies conducted through studying the disease transmission dynamics modeled by its compartmental transitions. We conjecture that the SEIR model can be improved by reducing the modeling assumptions implicit in its definition such as the explicit separation of the recovered individuals from the fatalities out of all those in the final 'removed' compartment. In the framework offered by \texttt{Pyro}, it is straightforward to modify the compartmental model structure which makes it extremely useful as a toolkit for practitioners interested in an experimental perspective on epidemiological modeling.

\begin{figure*}[ht]
	\begin{subfigure}{\textwidth}
		\centering
		% include first image
		\includegraphics[width=13cm, height=3.5cm]{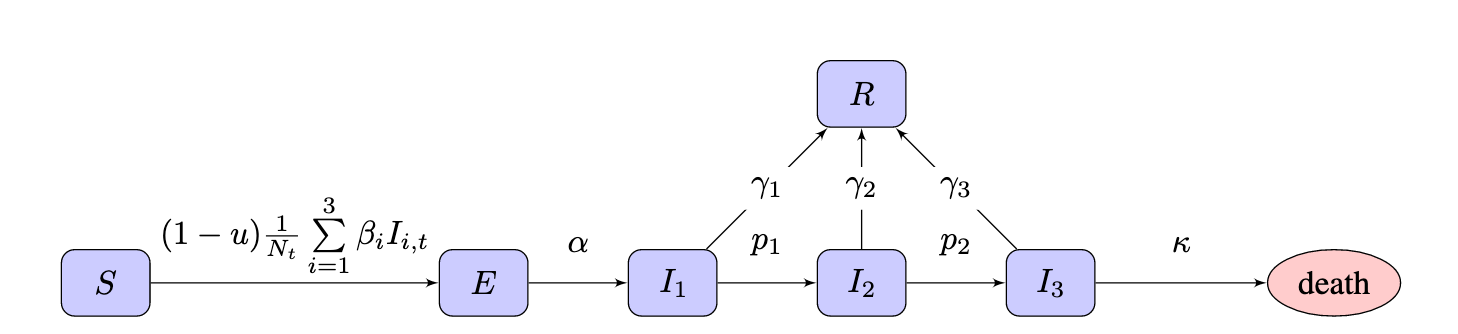}
	\end{subfigure}
	\caption{The SEI3RD Model from \cite{wood2020planning} along with our policy intervention parameter $u$}
	\label{fig:sei3rd_model}
\end{figure*}

As a template, we consider the idea of three-layered infection states corresponding to increasingly infectious 'spreaders' motivated by the need to tie epidemiology with medical physiology through modeling causal dynamic processes in time \cite{winters2020novel} for defining an SEI3RD model as shown in \ref{fig:sei3rd_model}.

For researchers, an excellent approach to reproducibly expand upon the existing literature could be following the approach of  \cite{ndairou2020mathematical}. They examine an eight-compartment model to obtain disease parameter estimates of a COVID-19 variant in Wuhan, China. They then conduct a sensitivity analysis of their model to examine the variance with respect to each parameter and compare their results with a numerical simulation to examine its suitability. Our motivation with the SEI3RD model is, in a similar vein, to expand upon the line of work leading to the SEIR model but with the difference of offering an open-source template to introduce new variants of compartmental models that can differ regionally.

\begin{figure*}[ht]
	\begin{subfigure}{\textwidth}
		\centering
		% include first image
		\includegraphics[width=14cm, height=14cm]{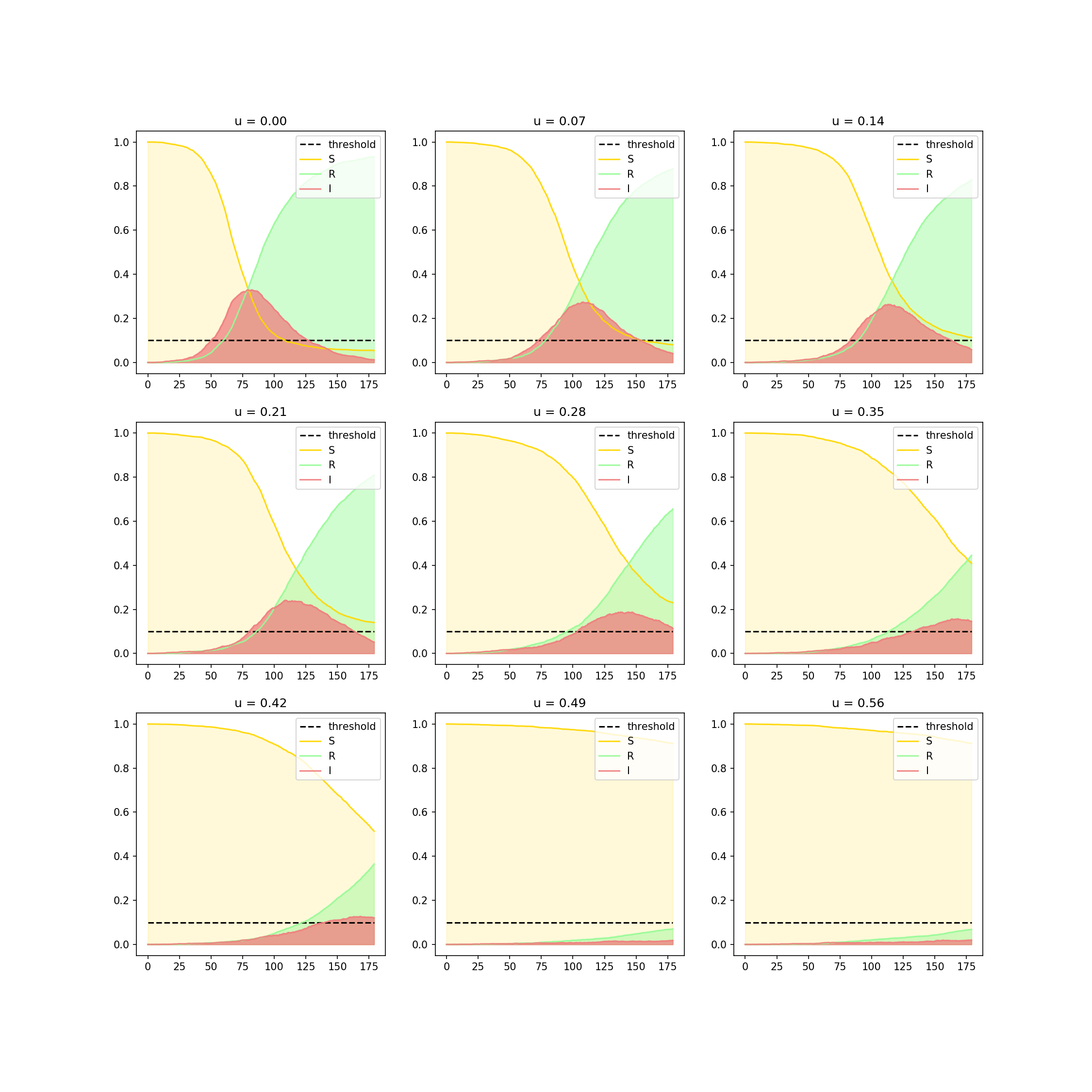}
	\end{subfigure}
	\caption{Visualizing the disease progression with time (along x-axis) and infected population fraction (along the y-axis) under varying levels of the policy intervention parameter $u$ for SIR.}
	\label{fig:sir_intervention}
\end{figure*}

\begin{figure*}[ht]
	\begin{subfigure}{\textwidth}
		\centering
		% include first image
		\includegraphics[width=11cm, height=11cm]{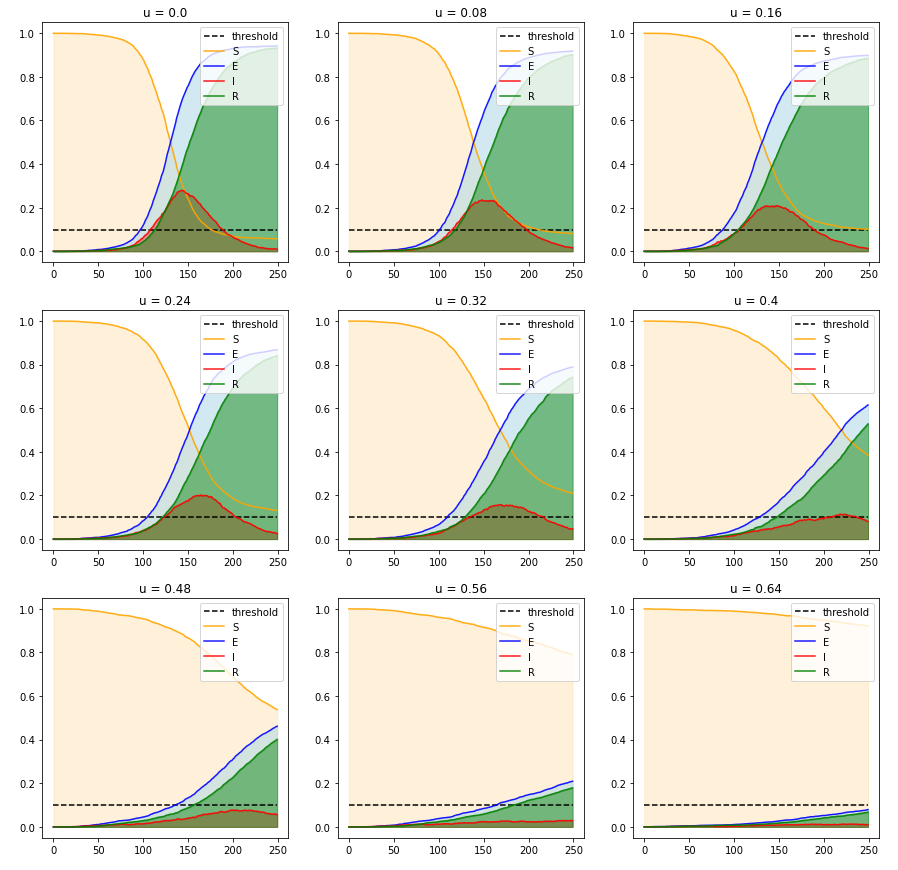}
	\end{subfigure}
	\caption{Visualizing the disease progression with time (along x-axis) and infected population fraction (along the y-axis) under varying levels of the policy intervention parameter $u$ for the SEIR model.}
	\label{fig:seir_intervention}
\end{figure*}

\subsection{Policy Interventions}
\label{section:policy_interventions}

The use of surgical face masks and face shields by healthcare and non-healthcare workers alike has been shown to significantly reduce or prevent the transmission of human coronaviruses and influenza viruses through respiratory droplets from symptomatic individuals in confined spaces \cite{leung2020respiratory, liang2020efficacy, chu2020physical}. This is an example of a type of intervention that falls into the class of non-pharmaceutical interventions (NPIs) which are important to model in light of the time it requires to implement substantial pharmaceutical interventions (PIs) such as vaccines. Our simulation, therefore, focuses on the modeling of NPIs, or what we term policy interventions, to allow us to build world-models that are reflective of disease spread in the absence of effective PIs. While vaccines may be the most effective long-term solution, there is significant impact of short term NPIs such as isolation and contact tracing \cite{grimm2021extensions, bertozzi2020challenges}. For this reason, our work focuses extensively on a framework to explore intervention strategies without the need for human-supervised search.

\begin{figure*}[ht]
	\begin{subfigure}{\textwidth}
		\centering
		% include first image
		\includegraphics[width=12cm, height=6cm]{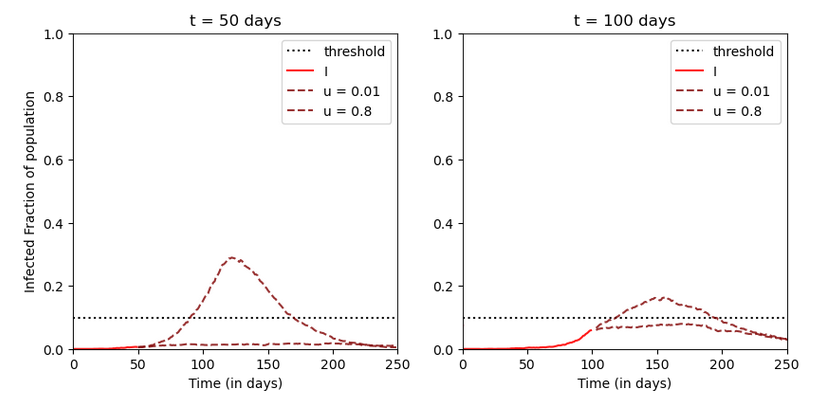}
	\end{subfigure}
	\caption{Adaptively Changing the Policy Intervention}
	\label{fig:adaptive_u}
\end{figure*}

When governments deal with diseases, they may take certain measures that result in limiting the exposure of the population to the disease. These measures may range from mild as in washing hands to stringent as in enforcing a complete lockdown \cite{usagov21, imf21}. While some lines of work focus on how to monitor their impact \cite{giudici2020monitoring, vasconcelos2020modelling}, we propose an algorithm to explore new strategies for control. The goal of epidemiological modeling, particularly the spread of pandemics is to be able to infer the actions necessary to limit their spread. Every policy to address this is designed to intervene on the rate of spread through natural or artificial means, for a short or long term duration. However, most modeling approaches treat it as affecting the transmission in a deterministic manner whereas in reality, the impact of government policies varies with time. For this reason, we decouple modeling of the policy intervention $u$ from the transmission probability $\beta$. This allows us to introduce our greedy search algorithm described in \label{adaptive_algo} (also see \ref{fig:adaptive_u}).

% \begin{figure}
%     \centering
    
    \begin{algorithm}
    \SetAlgoLined
    \KwResult{Sequence of policy interventions}
     initialize sequence\;
     \While{$t \leq T$}{
      simulate compartmental transitions\;
      \For{$u = u_i$, increasing from $0$ to $1$}{
           simulate a trajectory with $R_0$, $\rho$\;
           \If{infections $\leq$ threshold}{
               add $u_i$ to sequence\;
               break\;
           }
        }
     }
    \caption{\label{fig:adaptive_algo} Adaptive Interventions through Greedy Search}
    \end{algorithm}

Modelling policy interventions for COVID-19 has been the focus of a large body of work (\cite{giudici2020monitoring, vasconcelos2020modelling, giordano2020modelling, chen2020modeling}). In our work, we start with the similar implementation of a policy intervention defined by the parameter $u$ which modifies the SIR compartmental model as per the equation below:

\begin{align*}
    \beta_1 &= (1 - u) \beta \\
    \beta &= \beta_1 \\
\end{align*}

Recall that $\beta$ was the transmission probability of an individual from the pool of susceptible individuals to the next compartment which is model-dependent. Changing the values of this policy intervention parameter manually, as in $u$ taking on a deterministic set of values, and simulating the resulting trajectories results in the plot shown in \ref{fig:sir_intervention} for the SIR model and \ref{fig:seir_intervention} for the SEIR model. The horizontal dashed line indicates the $10\%$ threshold of infections. This approach allows us to monitor what could have been the impact of governmental policies such as a lockdown implemented over a long time period. However, $u$ may not be the same at all points in time. Thus, we introduce the algorithm \ref{adaptive_algo} to consider a sequence of policy interventions that is analogous to periodic policy updates or a variance in the impact of the same policies. A visual analogy of the algorithm is offered by illustrating the best and worst case outcomes of selecting different policy interventions at two time-steps in \ref{fig:adaptive_u} with a description below.

The current fraction of infected population is represented in bright red as a solid line. At each timestep, we perform a simulation of possible trajectories for different values of the policy intervention $u$. For example, at a time $t = 50$ days we simulate the possible trajectories (dashed lines). The plots show the best and worst-case options for the progress of the disease depending on the magnitude of the policy intervention parameter $u$. This is repeated at each time step $t$, with the illustration of $t=100$ offered in a separate plot.

A greedy solution, as demonstrated in the algorithm \ref{adaptive_algo}, is to pick the $u$ in a short-term optimal manner such that it keeps the infected fraction just within the threshold. However, this might still result in a sequence of large interventions for a highly infectious disease (long-term lockdowns). There are better strategies that enact more stringent measures early on so that the overall impact of interventions across the time series is not as high. Alternatively, we might weight the outcomes of infected individuals exceeding a threshold by the amount of time it would take to breach the threshold and thereby determine a reasonable intervention to perform. There are different ways of picking the optimal policy intervention depending on the choice of utility function and optimisation technique, including neural networks. We believe that the choice should vary subject to demographic factors, effectiveness of government policies, durations for which governments can implement stringent measures and acceptable thresholds of infected individuals.

\subsection{Limitations}

We study the spread of infectious diseases in terms of the COVID-19 pandemic in an attempt to unify the different models to highlight the utility of PPLs. For this reason, while we do study the predictions of our models by evaluating the parameter estimates against those described in the existing literature \cite{verity2020estimates, hoseinpour2020understanding, li2020substantial, kucharski2020early, wang2020evolving} in some experiments, it is more interesting for us to focus, in this work, on our technique and its extensions than the fact that we achieve similar results to those in related literature. We provide a complete comparison of some region-wise parameter estimates with real-world data in \ref{table:covid_R0} but emphasize that the novelty is in the ease of application of this technique across the table, its adaptability to incorporate modeling assumptions, and flexibility of inference across a spectrum of compartmental models.

It is challenging to model a disease with convenient mathematical assumptions implicit in compartmental models, that largely ignore individual-level differences. While there are some inferences that are plausible to make, it is clear that we must not jump to immediate conclusions in disease modeling due to the fact that there is almost always non-trivial uncertainty associated with parameter estimates in that multiple models might reflect similar initial trajectories \cite{atkeson2020deadly}. Furthermore, certain types of interventions might not be as effective as others \cite{manchein2020strong}. However, modeling interventions is one way of translating theory into practical advice for making policy decisions \cite{thompson2020epidemiological} and it is promising that SEIR models perform well over time in multiple demographies with regards to predicting the spread of the disease \cite{atkeson2021parsimonious}. This strengthens the argument that we would like to further reduce our modeling assumptions and therefore, uncertainty in parameter estimates, in order to make more confident policy recommendations eventually. This motivates the use of the SEI3RD model (among other variants in \nameref{section:extensions}) which is more powerful than the class of SEIR models since it is a derivative that allows for more granular transmission dynamics between compartments. However, even with the SEI3RD model, we need to add in external interventions, consider jointly conditioning on observed fatalities, and inductive biases on transition probabilities based on estimated incubation period, infection severity, stratification by demographic factors, and consider utility 

\end{document}